%% file: main.tex
\documentclass{article}

\usepackage[round]{natbib}
\usepackage[hyphens]{url}            
\usepackage[colorlinks=true,citecolor=orange,linkcolor=blue,breaklinks=true]{hyperref}       

\usepackage{amssymb,amsmath}
\usepackage{amsthm}
\usepackage{cleveref}
\usepackage{bm}
\usepackage{mathtools}
\usepackage{todonotes}
\usepackage{booktabs}

\newcommand{\E}{\mathbb{E}}
\newcommand{\err}{\mathcal{E}}
\newcommand{\emperr}{\widehat{\mathcal{E}}}
\newcommand{\vecemperr}{\bm{\widehat{\mathcal{E}}}}
\newcommand{\vecerr}{\bm{\mathcal{E}}}
\newcommand{\emprisk}{\widehat{\mathcal{L}}}
\newcommand{\risk}{\mathcal{L}}
\newcommand{\probmeasure}{\mathcal{M}^+_1}
\renewcommand{\Pr}{\mathbb{P}}
\renewcommand{\Re}{\mathbb{R}}
\newcommand{\KL}{\operatorname{KL}}
\newcommand{\bigO}{\mathcal{O}}
\newcommand{\kl}{\operatorname{kl}}
\renewcommand{\vec}[1]{\bm{#1}}
\newcommand{\1}{\bm{1}}
\newcommand{\exloss}{\tilde{\ell}}
\newcommand{\comp}{\textsc{kl}}
\newcommand{\lgdelta}{\textsc{lg}}
\newcommand{\logisticregression}{\operatorname{LGR}}
\DeclarePairedDelimiter\ceil{\lceil}{\rceil}

\newtheorem{proposition}{Proposition}
\newtheorem{theorem}{Theorem}

\allowdisplaybreaks

\usepackage[
    verbose=true,
    letterpaper,
    textheight=9in,
    textwidth=5.5in,
    top=1in,
    headheight=12pt,
    headsep=25pt,
    footskip=30pt]{geometry}

\makeatletter
    \def\@maketitle{%
  \newpage
  \null
  \vskip 2em%
  \begin{center}%
  \let \footnote \thanks
    {\LARGE \@title \par}%
    \vskip 1.5em%
    {\small
      \lineskip .5em%
      \begin{tabular}[t]{c}%
        \@author
      \end{tabular}\par}%
  \end{center}%
  \par
  \vskip 1.5em}
\makeatother

\title{Tighter PAC-Bayes Generalisation Bounds by Leveraging Example Difficulty}

\author{%
  Felix Biggs \\
  Department of Computer Science\\
  University College London and Inria\\
  London \\
  \texttt{ucabbig@ucl.ac.uk} \\
  \and
  Benjamin Guedj \\
  Department of Computer Science\\
  University College London and Inria\\
  London \\
  \texttt{b.guedj@ucl.ac.uk} \\
}

\begin{document}

\maketitle

\begin{abstract}
  We introduce a modified version of the excess risk, which can be used to obtain tighter, fast-rate PAC-Bayesian generalisation bounds.
  This modified excess risk leverages information about the relative hardness of data examples to reduce the variance of its empirical counterpart, tightening the bound.
  We combine this with a new bound for \([-1, 1]\)-valued (and potentially non-independent) signed losses, which is more favourable when they empirically have low variance around \(0\).
  The primary new technical tool is a novel result for sequences of interdependent random vectors which may be of independent interest.
  We empirically evaluate these new bounds on a number of real-world datasets.
\end{abstract}

\section{\sloppy INTRODUCTION AND OVERVIEW}

Generalisation bounds are of paramount importance in machine learning, both for understanding generalisation, and for obtaining guarantees for predictors.
Obtaining the tightest possible bounds shines light on the former and leads to numerically better guarantees for the latter.

Consider a parameterised learning problem where we are interested in training a predictor $h_w$ depending on weights $w$ (\emph{e.g.}, a neural network).
In PAC-Bayes, predictions are typically made by drawing randomised weights \(W \sim \rho\) where $\rho$ is a so-called posterior distribution, then predicting \(h_{W}(x)\) for some input $x$.
Thus the learning is moved from the parameter \(w\) to a distribution \(\rho\) over \(W\).

PAC-Bayesian generalisation bounds \citep{STW1997,McAllester1998,McAllester1999,catoni2007} allow for quantifying the generalisation performance of predictors of the form $h_W$ with high probability.
They can also be used as a stepping stone to proving bounds where \(w\) is not random, for example for majority votes \citep{DBLP:conf/nips/MasegosaLIS20,zantedeschi2021,DBLP:journals/corr/abs-2206-04607}.
The recent surge in attention given to the PAC-Bayesian approach partially derives from a number of works establishing numerically non-vacuous bounds for neural networks with randomised \citep{dziugaite2017computing,NIPS2018_8063,zhou2018nonvacuous,NIPS2019_8911,DBLP:journals/entropy/BiggsG21,DBLP:journals/corr/abs-2006-10929,JMLR:v22:20-879} or non-randomised \citep{biggs2022shallow} weights on real-world datasets.
We refer to \citet{guedj2019primer} and \citet{alquier2021userfriendly} and the many references therein for a broad introduction to PAC-Bayes.

Two terms commonly appear in PAC-Bayes bounds: \(\comp := \KL(\rho, \pi)\), which defines the complexity of \(\rho\) as a Kullback-Leibler divergence from some sample-independent reference measure (usually referred to as ``prior'') \(\pi\); and \(\lgdelta\), a term logarithmic in the probability \(\delta\).
If the number of examples is \(m\), then at worst \(\lgdelta \le \bigO(\log(m/\delta))\).
The simplest such bound for bounded losses \citep{McAllester1998} takes the form
\[
  \text{generalisation risk of } \rho \le \bigO\left( \sqrt{\frac{\comp + \lgdelta}{m}} \right),
\]
holding with probability at least \(1{-}\delta\) over the sample.
The above is rarely tight, and was greatly improved by the bound of \citet{maurer}, which we discuss further in \Cref{section:kl-bounds}.
Maurer's bound has the advantage that it can (when the empirical loss of \(\rho\) is small) achieve a faster rate of convergence, where the dependence \(\bigO(\sqrt{\comp/m})\) is improved to the ``fast-rate'' \(\bigO(\comp/m)\).
Since commonly \(\comp \gg \lgdelta\), this can lead numerically to much tighter bounds.

A major question in (PAC-Bayesian) learning theory is \emph{under what conditions such rates can be possible}.

As in VC theory, such fast-rates are possible when the empirical risk of \(\rho\) is zero, but it is also possible to get close to this fast regime under more general conditions.
Getting such faster rates is a primary motivation for ``Bernstein'' and ``Bennett''-type bounds (which leverage low variance to get faster rates) in classical learning theory, as well as for the introduction of the excess loss, which combines nicely with the former.

\subsection{NOTATION}

In order to further discuss existing approaches, we define our terms more thoroughly.
In the following, we examine different PAC-Bayesian generalisation bounds for bounded losses \(\ell: \mathcal{W} \times \mathcal{Z} \to [0, 1]\) (where the specific range \([0, 1]\) is w.l.o.g. due to the possibility of rescaling).
We let \(\mathcal{W}\) denote the weight space and \(\mathcal{Z}\) is the sample space.

A generalisation bound is an upper bound on the risk\footnote{extended by abuse of notation in a PAC-Bayesian setting to \(\risk(\rho) := \E_{W \sim \rho} \E_{Z \sim \mathcal{D}} \ell(W, Z)\).} \(\risk(w) := \E_{Z \sim \mathcal{D}} \ell(w, Z)\), that holds for some data-dependent hypothesis\footnote{or PAC-Bayesian posterior distribution \(\rho \in \probmeasure(\mathcal{W})\).} \(w\).

The excess risk is introduced by comparing the loss of our hypothesis \(w\) to a fixed ``good'' hypothesis \(w^{\star}\) (we leave aside for now the question of choosing \(w^{\star}\)) in a modified loss function, \(\exloss(w, z) = \ell(w, z) - \ell(w^{\star}, z).\)
This has the population and sample counterparts \[\err(w) := \E_{Z \sim \mathcal{D}}\exloss(w, Z) = \risk(w) - \risk(w^{\star})\] and \[\emperr(w) := \E_{Z \sim \operatorname{Uniform}(S)}\exloss(w, Z).\]

\paragraph{Our contributions.}
From the above starting point, we pursue two new parallel and complimentary directions of improvement.
The aim of these ideas is to show how PAC-Bayes bounds can be made tighter by using information from the training set more efficiently.
Firstly, we prove a new and tighter bound on the excess loss.
This can be used to prove new generalisation bounds which also attain faster rates under slightly different conditions from Maurer's bound.
We then go on to provide a generalisation of the excess risk which allows \(w^{\star}\) to be learned from the stream of data as we receive it.

\subsection{FAST RATES AND EXCESS LOSSES}

The simplest PAC-Bayesian bound which can achieve fast rates (and therefore tighter bound values) is the following:
\begin{equation}\label{eq:maurer-relax}
  \risk - \emprisk \,\le\, \sqrt{\frac{\comp + \lgdelta}{m} \cdot 2\emprisk} \,+\, 2 \frac{\comp + \lgdelta}{m}.
\end{equation}
This bound\footnote{Note that for the sake of clarity and without loss of generality, we will make the slight notational abuse of omitting the argument of $\risk$, $\emprisk$, $\err$ and $\emperr$ when the (PAC-Bayes) context is clear.} (which is a relaxation of Maurer's bound, see \Cref{section:kl-bounds}) has a well-studied form common in classical learning theory where the \(\comp\) term is replaced by a different complexity term.
When \(\emprisk \to 0\) it achieves the fast rate on \(\risk - \emprisk\) of \(\bigO(\comp / m)\) and will be numerically tighter, but otherwise (for example, on a difficult dataset where \(\emprisk\) is large) the square root term typically dominates.

A common question in learning theory has therefore been on whether empirical risk under the square root can be replaced by something faster-decaying, like a variance \citep{DBLP:conf/nips/TolstikhinS13} or an \emph{excess} risk.
For example, \citet{unexpected} prove the ``Unexpected Bernstein'' PAC-Bayes bound
\[
  \err \le \emperr + \bigO \left(\sqrt{\frac{\comp + \lgdelta}{m} \cdot \widehat{V}} + \frac{\comp + \lgdelta}{m} \right),
\]
where \(\widehat{V}(\rho) = \E_{W \sim \rho} [\frac{1}{m}\sum_{i=1}^m |\ell(W, Z_i) - \ell(w^{\star}, Z_i)|^2] \le \emperr(\rho)\).
The idea is that the second loss term in the excess risk ``de-biases'' and reduces the variance, so that if the predictors err on a similar set of examples, \(\emperr(w)\) will be small, giving a faster rate.
Such bounds on \(\err\) can be converted back into generalisation bounds, by using that \(\risk(w^{\star}) - \emprisk(w^{\star}) \le \bigO(\sqrt{\lgdelta / m})\) (since \(w^{\star}\) is independent of the dataset) to get a bound like
\begin{equation}\label{eq:relaxed-general-form}
  \risk \le \emprisk + \bigO \left(\sqrt{\frac{\comp + \lgdelta}{m} \cdot \widehat{V}} + \frac{\comp + \lgdelta}{m} + \sqrt{\frac{\lgdelta}{m}} \right).
\end{equation}
Since in most cases \(\lgdelta \ll \comp\), the final term is usually an insignificant price compared with the reduction from \(\emperr\) to \(\widehat{V}\).
The rate of the final term can also be improved even further using assumptions about the noise (as examined at length in \citealp{unexpected}), or using dataset evaluations of the loss of \(w^{\star}\).

A problem with this approach is the fact that \(w^{\star}\) must be independent of the data. This means we must split the dataset as with PAC-Bayes data-dependent priors \citep[as in, \emph{e.g.},][]{DBLP:journals/jmlr/Parrado-HernandezASS12,DBLP:conf/nips/RivasplataSSPS18,unexpected,perezortiz2021learning}, into parts used to produce \(w^{\star}\) (and potentially learn a prior), and to actually apply the bound to.
This reduces the effective sample size in the bound (\emph{e.g.}, from \(m\) to \(m/2\) when a 50-50 split is used).
This issue can be partially circumvented through the use of forwards-backwards ``informed'' priors, but in expectation over different splits of the data this approach is actually weaker than the naive splitting procedure.

\subsection{KL-BASED BOUNDS}\label{section:kl-bounds}

The most well known (and often tightest) PAC-Bayesian bound for bounded losses \(\in [0, 1]\) is Maurer's bound \citep{maurer}:
\[
  \kl\left(\emprisk(\rho) \| \risk(\rho)\right) \le \frac{\KL(\rho, \pi) + \log\frac{2\sqrt{m}}{\delta}}{m}
\]
where \(\kl(q \| p) = q \log\frac{q}{p} + (1-q) \log\frac{1-q}{1-p}\) is the KL divergence between Bernoulli distributions of biases \(q\), \(p\).
This bound can be inverted to obtain an upper bound directly on \(\risk\) by defining the inverse \[\kl^{-1}(u \| b) := \sup \{r : \kl(u \| r) \le b\}.\]
The bound in \Cref{eq:maurer-relax} is obtained through the relaxation \(\kl^{-1}(u \| b) \le u + \sqrt{2bu} + 2b\) \citep{DBLP:conf/colt/McAllester03}.
However, note that this lower bound can be considerably weaker, as it does not leverage the combinatorial power of the small-kl.

We note also that although the small-kl bound can be re-scaled to use the excess loss, this leads to a bound like \Cref{eq:relaxed-general-form} with \(\widehat{V} = \frac{\emprisk + 1}{2} \ge \frac12\), which does not lead to fast rates.

Recently, \citet{adams} proved a generalisation of this bound which holds for \emph{vector}-valued losses, \(\bm{\ell}: \mathcal{W} \times \mathcal{Z} \to \Delta_M\) (with \(\Delta_M\) the M-dimensional simplex),
\[
  \kl\left(\widehat{\vec{U}} \| \vec{\mu} \right) \le \frac{\KL(\rho, \pi) + \log\frac{\xi(M, m)}{\delta}}{m}.
\]
where \(\widehat{U}_i(w) = \frac{1}{m} \sum_{i=1}^m \ell_i(w, z)\) and \(\vec{\mu} = \E_S \widehat{\vec{U}}\), and \(\xi(M, m)\) is a polynomial function of \(m\).
Inverting such a bound is somewhat more complicated, but we can use it to obtain an upper bound on \(\sum_i \alpha_i U_i\) for some set of coefficients \(\alpha_i\).
This is the tool we will use to obtain our bounds for signed losses.

\paragraph{Some more of our contribution.}
Firstly, we observe that \Cref{eq:maurer-relax} is a \emph{relaxation} of Maurer's bound, and this weakening leads to a loss of some of the tightness of the original.
We give a new bound which leverages the tightness of kl-based bounds like Maurer's, but also relaxes to a form like that in \Cref{eq:relaxed-general-form}.
Specifically, it reduces to the form given in \Cref{eq:relaxed-general-form} with \(\widehat{V}(\rho) = \E_{W \sim \rho}\frac{1}{m} |\ell(W, Z_i) - \ell(w^{\star}, Z_i)|\).
This is very similar (if slightly larger) to the term given in \citet{unexpected}, although both are equivalent when \(\ell \in \{0, 1\}\), as for example with the misclassification loss.
However this form of our bound is only a relaxation, and the kl-type formulation that we give for it is considerably tighter.

\section{WARM UP}

In this section we give a simplified version of our main results, discussing only classification.
In this setting, \(\mathcal{Z} = \mathcal{X} \times \mathcal{Y}\) for \(\mathcal{Y} = \{1, \dots, c\}\), and \(S = \{(X_i, Y_i)\}_{i=1}^m\).
Our predictions are given by \(h_w(x)\) for \(w \in \mathcal{W}\) and we consider the misclassification loss, \(\ell(w, (x, y)) = \1_{h_w(x) \ne y}\).

When considering the excess loss, there are effectively two different error types:
\begin{align*}
    \emperr_{+} &= \E_{W \sim \rho} \left[\frac{|\{ (x, y) \in S : h_W(x) \ne y, h_{w^{\star}}(x) = y\}|}{m} \right],\\
    \emperr_{-} &= \E_{W \sim \rho} \left[\frac{|\{ (x, y) \in S : h_W(x) = y, h_{w^{\star}}(x) \ne y\}|}{m}\right].
\end{align*}
Thus we are merely counting the numbers of two different types of loss: an error using \(w\) but not using \(w^{\star}\), and the converse.
If neither predictor or both predictors err, this incurs no loss.
These two error types have simple interpretations as counts, which is similar to the work of \citet{adams}.
Collecting the error type counts into the vector
\[\vecemperr(\rho) = [\emperr_+(\rho), \emperr_-(\rho), 1 - \emperr_+(\rho) - \emperr_-(\rho)]^T,\]
the bound from \citet{adams} can be used to bound \(\bm{\ell}\), and inverted with \(\vec{\alpha} = [1, -1, 0]\) to upper bound the excess loss.
This shows that
\[
    \err(\rho) \le \phi\left( \vecemperr(\rho), \; \frac{\KL(\rho, \pi) + \log\frac{2m}{\delta}}{m} \right)
\]
where \(\phi(\vec{u}, b) := \sup \left\{ r_1 - r_2 : \vec{r} \in \Delta_3, \, \kl(\vec{u} \| \vec{r} ) \le b \right\}\).

This form leverages the tightness of the kl bound and we use it in practice, but for the sake of intuition we show that this also implies the weaker relaxed form
\begin{equation}\label{eq:relaxed-ours}
    \err(\rho) \le \emperr(\rho) + 2\sqrt{\frac{\comp + \lgdelta}{m} \cdot (\emperr_+ + \emperr_-)} + 2\frac{\comp + \lgdelta}{m}
\end{equation}
with \(\lgdelta = \log(2m/\delta)\).
The square root term can alternatively be written as
\[
  \emperr_+ + \emperr_- = \frac{1}{m} \sum_{i=1}^m \Pr_{W \sim \rho}(h_W(X_i) \ne h_{w^{\star}}(X_i)),
\]
a form which commonly appears in learning theory.

This result can be combined with a (test set) bound on \(\risk(w^{\star}) - \emprisk(w^{\star})\) to provide a generalisation bound for \(\rho\), as in \Cref{eq:relaxed-general-form}.

For intuition, we point out that this basic result (though not the relaxation) can be proved straightforwardly by application of the results of \citet{adams} to a vector valued loss, \(\vec{\ell}: \mathcal{W} \times \mathcal{Z} \to \Delta_3\), where
\begin{align*}
\ell_1 &= \1_{h_w \ne y, h_{w^{\star}} = y},  \\
\ell_2 &= \1_{h_w = y, h_{w^{\star}} \ne y}, \\
\ell_3 &= 1 - \ell_1 - \ell_2.
\end{align*}
Our main result generalises the above in two different directions.
We adapt the bound to work for any bounded loss function, and we generalise the excess loss to use a richer and more data-informed de-biasing process.
In particular, we show how the first \(i-1\) examples can be used to learn a \(w^{\star}_i\) which is used to de-bias the loss incurred by example \(i\).
If the procedure used to learn the \(w^{\star}_i\) is similar to that used to learn \(\rho\), and is relatively stable to changes in dataset size, the errors of \(\rho\) and the \(w^{\star}_i\) will be highly correlated, reducing the excess risk and tightening the overall bound.
This approach can be easily generalised to stochastic algorithms, a procedure mentioned by \citet{unexpected} as ``online estimators'', but not used empirically.
We note that, unlike with data-dependent priors in PAC-Bayes bounds, no data-splitting is necessary for this procedure, the bound still uses all of the training data with \(m = |S|\).

\section{MAIN RESULTS}

Here we consider a general setting that allows easy derivation of results about the (generalised) excess risk.
We will use these to obtain faster rates for the excess risk in cases where the standard risk bounds converge more slowly.

We introduce notation for a sequence of examples, \(z_{1:i} = (z_1, \dots, z_i) \in \mathcal{Z}^{\star}\), where \(A^{\star} := \emptyset \cup \bigcup_{i=1}^\infty A^i\) is the set of sequences of elements in set \(A\) and we notate \(z_{1:0} = \emptyset\).
In the following, \(\ell: \mathcal{W} \times \mathcal{Z} \to [0, 1]\) is a bounded loss and \(\ell^{\star}: \mathcal{Z}^{\star} \to [0, 1]\) is a bounded de-biasing function.
After choosing the de-biasing function, our loss for each example will take the form
\[
  \tilde{\ell}(w, Z_{1:i}) := \ell(w, Z_i) - \ell^{\star}(Z_{1:i}).
\]

Given a sample \(S = Z_{1:m}\) of i.i.d. random variables \(Z_i\) drawn from an unknown data-generating distribution \(\mathcal{D} \in \probmeasure(\mathcal{Z})\), and a PAC-Bayesian posterior \(\rho \in \probmeasure(\mathcal{W})\), we define the generalised excess risk (with respect to \(\ell^{\star}\)) as
\[
  \err(\rho) :=  \E_{W \sim \rho}\E_Z[\ell(W, Z)] - \frac{1}{m} \sum_{i=1}^m \E_{Z_i}[\ell^{\star}(Z_{1:i})|Z_{1:i-1}].
\]
We note this is now actually a random variable dependent on \(Z_{1:m-1} = S \setminus Z_m\) through the de-biasing term.
We want an upper bound on \(\err(\rho)\) holding with high probability over \(S\), which should be simultaneously true for all posteriors \(\rho\) given a fixed (sample-independent) reference measure (a.k.a. PAC-Bayesian prior) \(\pi \in \probmeasure(\mathcal{W})\).

In order to do so, we distinguish two different types of sample errors (corresponding to positive and negative parts) with corresponding generalised empirical risk values:
\begin{align*}
    \emperr_{+}(\rho) &:= \E_{W \sim \rho} \left[\frac{1}{m} \sum_{i=1}^m \max(\tilde{\ell}(W, Z_{1:i}), 0) \right], \\
    \emperr_{-}(\rho) &:= \E_{W \sim \rho} \left[\frac{1}{m} \sum_{i=1}^m \max(-\tilde{\ell}(W, Z_{1:i}), 0) \right].
\end{align*}
These correspond to error types when \(\exloss > \ell^{\star}\) and \(\exloss < \ell^{\star}\) respectively.
For notational convenience we collect these in the vector \[\vecemperr(\rho) = [\emperr_+(\rho), \emperr_-(\rho), 1 - \emperr_+(\rho) - \emperr_-(\rho)]^T.\]
We also define the expected formulations of the above as
\begin{align*}
    \err_{+} &:= \E_{W \sim \rho} \left[\frac{1}{m} \sum_{i=1}^m \E_{Z_i}[\max(\tilde{\ell}(W, Z_{1:i}), 0)|Z_{1:i-1}] \right] \\
    \err_{-} &:= \E_{W \sim \rho} \left[\frac{1}{m} \sum_{i=1}^m \E_{Z_i}[\max(-\tilde{\ell}(W, Z_{1:i}), 0)|Z_{1:i-1}] \right].
\end{align*}
also collected into vector \[\vecerr(\rho) = [\err_+(\rho), \err_-(\rho), 1 - \err_+(\rho) - \err_-(\rho)]^T.\]

\begin{theorem}\label{th:main-bound}
  For any measurable \(\ell\) and \(\ell^{\star}\) as defined above and any \(\mathcal{D} \in \probmeasure(\mathcal{Z})\),
  with probability at least \(1{-}\delta\) over \(S \sim \mathcal{D}^m\) simultaneously for all \(\rho \in \probmeasure(\mathcal{W})\),
  \[
      \kl \left( \vecemperr(\rho) \middle\| \vecerr(\rho) \right) \le \frac{\KL(\rho, \pi) + \log\frac{2m}{\delta}}{m}.
  \]
  Here \(\kl(\vec{u} \| \vec{v}) = \sum_i u_i \log \frac{u_i}{v_i}\) is the KL divergence between categorical variables with parameters \(\vec{u}\) and \(\vec{v}\).
  This is inverted to obtain that
  \[
      \err(\rho) \le \phi\left( \vecemperr(\rho), \; \frac{\KL(\rho, \pi) + \log\frac{2m}{\delta}}{m} \right),
  \]
  where \(\phi(\vec{u}, b) := \sup \left\{ r_1 - r_2 : \vec{r} \in \Delta_3, \, \kl(\vec{u} \| \vec{r} ) \le b \right\}\).
\end{theorem}

The relaxed form given by \Cref{eq:relaxed-ours} is still valid and gives intuition about the bound; in this more general case,
\[
  \emperr_+ + \emperr_- = \E_{W \sim \rho}\frac{1}{m} \sum_{i=1}^m |\ell(W, Z_i) - \ell^{\star}(Z_i)|.
\]
As mentioned above, this relaxed form is very similar to that of \citet{unexpected}, and in the case of the 0-1 misclassification loss the terms \(\widehat{V}\) are equivalent.

We examine instances of this bound in the next sections.
We also note here that it is possible to obtain gradients of \(\phi\) with respect to both of its arguments using a procedure outlined by \citet{adams}, which could be very useful in optimising the bound directly as an objective.

\subsection{REDUCTION TO BOUNDED LOSS}\label{section:relax-to-bounded}

In order to show that the new bound leverages the tightness of a small-kl-based bound as well as relaxing to a simple fast-rate form, we show that it can be used to recover (up to a factor $2$ in front of the logarithmic term) the non-relaxed version of Maurer's bound.
Setting \(\ell^{\star} = 0\) gives \(\emperr_- = 0\) and
\[ \emperr_+(\rho)= \E_{W \sim \rho} \left[ \frac{1}{m} \sum_{i=1}^m \ell(W, Z_i) \right] = \emprisk(\rho).\]
The following proposition is then used to reduce to the Bernoulli small-kl.

\begin{proposition}
  For \(u \in [0, 1]\) and \(b > 0\)
  \[ \phi([u, 0, 1-u], b) = \kl^{-1}(u \| b). \]
\end{proposition}

\begin{proof}
  We know that \(\kl([u, 0, 1-u] \| \vec{r}) \ge \kl(u \| r_1)\) by \citet[Proposition 9]{adams}, with equality when \(r_2 = 0\).
  Therefore we can set \(r_2 = 0\) without making it any more difficult to satisfy the constraint \(\kl([u, 0, 1-u] \| \vec{r}) \le b\).
  In this case \[\phi([u, 0, 1-u], b) = \sup \{r_1 : \kl(u \| r_1) \le b\} \]
  which is the definition of \(\kl^{-1}\).
\end{proof}

Based on this we (almost) recover Maurer's bound for the expected risk \(\risk(\rho) := \E_{S \sim \mathcal{D}^m}[\emprisk(\rho)] = \err(\rho)\) as
\[
    \risk(\rho) \le \kl^{-1}\left( \emprisk(\rho) \middle\|  \frac{\KL(\rho, \pi) + \log\frac{2m}{\delta}}{m} \right).
\]
This differs from Maurer's bound only in the worse constant (\(1\) instead of \(\frac12\)) for the \(\log m\) term.

\subsection{BASIC EXCESS LOSS}

Our bound can also be reduced to one for the standard excess loss instead of using our more sophisticated data-dependent de-biasing term.
This is done by setting \(\ell^{\star}(z_{1:i}) = \ell(w^{\star}, z_i)\) where \(w^{\star}\) is independent of the data \(S\).
Based on this we recover the standard definition of the excess loss, since
\[
  \err(\rho) = \risk(\rho) - \risk(w^{\star}).
\]
In order to turn this into a generalisation bound like \Cref{eq:relaxed-general-form}, we note the following bound, due to \citet{hoeffding63} and put in that particular form by \citet{DBLP:journals/corr/abs-2205-07880}.
\begin{theorem}[Chernoff-Hoeffding]
  For any \(\mathcal{D}\) and \(w^{\star}\) independent of the sample \(S \sim \mathcal{D}\), with probability at least \(1{-}\delta\) over \(S\),
  \[
    \risk(w^{\star}) \le \kl^{-1}\left( \emprisk(w^{\star}) \middle\| \frac{\log\frac{1}{\delta}}{m} \right).
  \]
\end{theorem}
This can be easily adapted to \(\risk(\rho^{\star})\) for any data-independent distribution \(\rho^{\star}\) by combination with Jensen's inequality.
By Pinsker's inequality it further implies that \[\risk(w^{\star}) - \emprisk(w^{\star}) \le \sqrt{\frac{\log(1/\delta)}{2m}},\] through which we can put our bound in the form \Cref{eq:relaxed-general-form}.
We note that in combining these results, a union bound must be used, so the overall probability is reduced to \(1{-}2\delta\) rather than \(1{-}\delta\).

\subsection{ONLINE DEBIASING}

In order to utilise our main bound to obtain generalisation bounds, it is necessary to obtain a bound on the term
\[ \frac{1}{m} \sum_{i=1}^m \E_{Z_i}[\ell^{\star}(Z_{1:i})|Z_{1:i-1}]\]
appearing in the generalised empirical loss, which can be difficult in general.

A choice that \emph{does} lead to an interesting and numerically calculable bound is setting \(\ell^{\star}(Z_{1:i}) = \ell(\mathcal{A}(Z_{1:i-1}), Z_i)\), where \(\mathcal{A}: \mathcal{Z}^{\star} \to \mathcal{W}\) is some algorithm.
The algorithm $\mathcal{A}$ can be anything but a natural choice is to choose a similar algorithm to that used to obtain our posterior.
This approach can be easily generalised to stochastic algorithms, \(\mathcal{A}: \mathcal{Z}^{\star} \to \probmeasure(\mathcal{W})\) with \[\ell^{\star}(Z_{1:i}) = \E_{W' \sim \mathcal{A}(Z_{1:i-1})} \ell(W', Z_i).\]

Thus if we run an algorithm on the first \(i-1\) examples we can legitimately use its error on \(Z_i\) to de-bias the loss of \(w\) on example \(Z_i\).
In this way we are using information about the relative difficulty of examples to de-bias our bound and make it tighter.
In order to do this we need to show that these terms can be numerically bounded, as we do in the next theorem.

\begin{theorem}\label{th:debias-bound}
  For a sample \(S = Z_{1:m}\) of i.i.d. variables, let \(\rho_i^{\star}, i = 1, \dots, m\) be a sequence of online estimators, where each depends only on the \(i-1\) examples \(Z_{1:i-1}\).
  Let
  \[
    \risk^{\star} := \frac{1}{m} \sum_{i=1}^m \E_{Z}\E_{W \sim \rho_i^{\star}}[\ell(W, Z)].
  \]
  With probability at least \(1{-}\delta\) over \(S\),
  \[
    \risk^{\star} \le \kl^{-1}\left( \frac{1}{m} \sum_{i=1}^m \E_{W \sim \rho_i^{\star}}[\ell(W, Z_i)] \middle\| \frac{\log\frac{1}{\delta}}{m} \right).
  \]
\end{theorem}
A numerically evaluable generalisation bound can therefore be provided by combining this result with \Cref{th:main-bound}.
The above result can also be combined with Pinsker's inequality as in the previous section, to obtain a form of the bound like \Cref{eq:relaxed-general-form}, but with the online de-biasing being used.

Finally we note that the form given with online estimators is not the most general to which the above theorem applies: any way in which we can use the first \(i-1\) examples to choose a debiasing function for \(Z_i\) will work, as long as this choice of function is independent of \(Z_i\).
Thus, more sophisticated procedures could be tried: for example a neural network could be trained on \(Z_{1:i-1}\) to predict the optimal de-biasing function \(f: \mathcal{Z} \to [0, 1]\) to be used on the next example.

\section{PROOFS AND COROLLARIES}

Firstly, we prove two theorems which generalise theorems of \citet{adams} to random variables with a dependence structure, using ideas from \citet{DBLP:conf/uai/SeldinLCSA12}.
These results may be of interest in their own right.

\begin{theorem}[Generalisation of Lemma 5 in \citealp{adams} and Lemma 1 in \citealp{DBLP:conf/uai/SeldinLCSA12}]\label{th:martingale-vector-convexity}
  Let \(\vec{U}_1, \dots, \vec{U}_m\) be a sequence of random vectors, each in \(\Delta_M\), such that \[\E[\vec{U}_i | \vec{U}_{1}, \dots, \vec{U}_{i-1}] = \vec{\mu}_i\] for \(i = 1, \dots, m\).
  Let \(\vec{V}_1, \dots, \vec{V}_m\) be independent \(\operatorname{Multinomial}(1, M, \vec{\mu}_i)\) random vectors such that \(\E\vec{V}_i = \vec{\mu}_i\).
  Then for any convex function \(f: \Delta_M^m \to \Re\):
  \[
    \E[f(\vec{U}_1, \dots, \vec{U}_m)] \le \E[f(\vec{V}_1, \dots, \vec{V}_m)].
  \]
\end{theorem}

\begin{proof}

  Let \(E_M\) denote the set of canonical (axis-aligned) \(M\)-dimensional basis vectors, for example \(E_3 = \{[1, 0, 0], [0, 1, 0], [0, 0, 1]\}\).
  We will denote typical members of this set by \(\vec{\eta}_i\), and tuples \(\vec{\eta}_{1:m} = (\vec{\eta}_1, \dots, \vec{\eta}_m) \in \Delta_M^m\).
  Firstly we show that the definitions in the theorem lead to a Martingale-type result:
  \begin{align*}
  &\E\left[ \prod_{i=1}^m \vec{U}_i \cdot \vec{\eta}_i \right] \\
    &=\E_{\vec{U_{1:m-1}}}\left[ \left(\prod_{i=1}^{m-1} \vec{U}_i \cdot \vec{\eta}_i \right) \E_{\vec{U}_m}\left[\vec{U}_m | \vec{U}_{1:m-1}\right] \cdot \vec{\eta}_m \right] \\
    &=\E_{\vec{U_{1:m-1}}}\left[ \left(\prod_{i=1}^{m-1} \vec{U}_i \cdot \vec{\eta}_i \right) \vec{\mu}_m \cdot \vec{\eta}_m \right] \\
    &=\prod_{i=1}^{m} \vec{\mu}_i \cdot \vec{\eta}_i.
  \end{align*}

  In \citet[proof of Lemma 5]{adams}, it is shown that for any convex function \(f: \Delta_M^m \to \Re\) and \(\vec{u}_{1:m} = (\vec{u}_1, \dots, \vec{u}_m) \in \Delta_M^m\),
  \[
      f(\vec{u}_{1:m}) \le \sum_{\vec{\eta}_{1:m} \in E_M^m} \left( \prod_{i=1}^m \vec{u}_i \cdot \vec{\eta}_i \right) \, f(\vec{\eta}_{1:m}).
  \]
  Applying this result to the random variables \(\vec{U}_{1:m}\) and combining with the Martingale-type result leads to the following:
  \begin{align*}
    \E&[f(\vec{U}_{1:m})] \\
    &\le \E\left[\sum_{\vec{\eta}_{1:m} \in E_M^m} \left( \prod_{i=1}^m \vec{U}_i \cdot \vec{\eta}_i \right) \, f(\vec{\eta}_{1:m}) \right] \\
    &= \sum_{\vec{\eta}_{1:m} \in E_M^m} \E\left[ \prod_{i=1}^m \vec{U}_i \cdot \vec{\eta}_i \right] \, f(\vec{\eta}_{1:m}) \\
    &= \sum_{\vec{\eta}_{1:m} \in E_M^m} \left(\prod_{i=1}^m \vec{\mu}_i \cdot \vec{\eta}_i  \right) \, f(\vec{\eta}_{1:m}) \\
    &= \sum_{\vec{\eta}_{1:m} \in E_M^m} \left(\prod_{i=1}^m \Pr(\vec{V}_i = \vec{\eta}_i)  \right) \, f(\vec{\eta}_{1:m}) \\
    &= \E f(\vec{V}_{1:m}).
  \end{align*}
  The final step in the proof followed via the definition of expectation w.r.t. the \(\vec{V}_i\).
\end{proof}

\begin{theorem}[Martingale PAC-Bayes for Vector KL.]\label{th:martingale-pac-bayes}
Let \(\vec{U}_1(w), \dots, \vec{U}_m(w)\) be a sequence of random vector valued functions, each in \(\Delta_M\), such that \[\E[\vec{U}_i(w) | \vec{U}_{1}(w), \dots, \vec{U}_{i-1}(w)] = \vec{\mu}_i(w)\] for \(i = 1, \dots, m\) and all \(w \in \mathcal{W}\).
Define \[\widehat{\vec{U}}(\rho) := \E_{W \sim \rho}\left[\frac{1}{m} \sum_{i=1}^m \vec{U}_i(W)\right]\] and \[\bar{\vec{\mu}}(\rho) := \E_{W \sim \rho}\left[\frac{1}{m} \sum_{i=1}^m \vec{\mu}_i(W)\right].\]
Then for fixed \(\pi \in \probmeasure(\mathcal{W}), \delta \in (0, 1)\), with probability at least \(1{-}\delta\) (over \(\{\vec{U}_i(w) : i \in 1, \dots, m : w \in \mathcal{W}\}\)), simultaneously for all \(\rho \in \probmeasure(\mathcal{W})\),
\[
  \kl \left( \widehat{\vec{U}}(\rho) \middle\| \bar{\vec{\mu}}(\rho) \right) \le \frac{\KL(\rho, \pi) + \log\frac{\xi(M, m)}{\delta}}{m}
\]
where \(\xi(M, m)\) is defined for \(m \ge M\) by
\[
  \sqrt{\pi} e^{1/12m} \left( \frac{m}{2} \right)^{\frac{M-1}{2}} \sum_{k=0}^{M-1} {M \choose k} \frac{1}{(m\pi)^{k/2} \Gamma \left( \frac{M-k}{2} \right)}.
\]
\end{theorem}

\begin{proof}
  The proof begins by a common pattern in PAC-Bayesian proofs \citep[see, \emph{e.g.},][]{guedj2019primer,alquier2021userfriendly,picard2022divergence}.
  By Jensen's inequality, the Donsker-Varadhan change-of-measure theorem, Markov's inequality and the independence of \(\pi\) from \(\{\vec{U}_i, \vec{\mu_i}\}\), the following holds with at least \(1{-}\delta\) for any \(\rho\):
  \begin{align*}
    &m \kl\left( \widehat{\vec{U}}(\rho) \middle\| \bar{\vec{\mu}}(\rho) \right) - \KL(\rho, \pi) \\
    &\le m \E_{W \sim \rho} \kl\left( \frac{1}{m} \sum_{i=1}^m \vec{U}_i(w) \middle\| \bar{\vec{\mu}}(w) \right)  - \KL(\rho, \pi) \\
    &\le \log \E_{W \sim \pi} \left[ e^{ m \kl\left( \frac{1}{m} \sum_{i=1}^m \vec{U}_i(w) \middle\| \bar{\vec{\mu}}(w) \right) } \right] \\
    &\le \log \frac{1}{\delta} \E_{\vec{U}_i}\E_{W \sim \pi} \left[ e^{ m \kl\left( \frac{1}{m} \sum_i \vec{U}_i(w) \middle\| \bar{\vec{\mu}}(w) \right) } \right] \\
    &\le \log \frac{1}{\delta} \E_{W \sim \pi} \E_{\vec{U}_i}\left[ e^{ m \kl\left( \frac{1}{m} \sum_i \vec{U}_i(w) \middle\| \bar{\vec{\mu}}(w) \right) } \right].
  \end{align*}
  By applying \Cref{th:martingale-vector-convexity} to the inner term we find that
  \begin{align*}
    \E_{\vec{U}_i}&\left[ e^{ m \kl\left( \frac{1}{m} \sum_i \vec{U}_i(w) \middle\| \bar{\vec{\mu}}(w) \right) } \right] \\
    &\le \E_{\vec{V}_i}\left[ e^{ m \kl\left( \frac{1}{m} \sum_i \vec{V}_i(w) \middle\| \bar{\vec{\mu}}(w) \right) } \right] \\
    &\le \E_{\bar{\vec{V}}}\left[ e^{ m \kl\left( \bar{\vec{V}} \middle\| \bar{\vec{\mu}}(w) \right) } \right],
  \end{align*}
  where \(\bar{\vec{V}} \sim \operatorname{Multinomial}(m, M, \bar{\vec{\mu}}(w))\).
  The latter step follows as the expectation of a convex sum of Multinomial variables is maximised by variables having the same constants, \(\vec{\mu}_i = \bar{\vec{\mu}}\) \citep{hoeffding1956distribution}.
  This final term is shown in Corollary 7 of \citet{adams} to be upper bounded by \(\xi(M, m)\) uniformly for all \(w\).
  We divide both sides by \(m\) to obtain the theorem statement.
\end{proof}

In showing the simpler form of our bound we also use the following.
\begin{proposition}\label{prop:xi-bound}
  For any \(m\ge 3\), \(\xi(3, m) \le 2m\).
\end{proposition}

\begin{proof}
  For \(M = 3\) the upper bound in \Cref{th:martingale-pac-bayes} evaluates to
  \[
    \frac12 e^{\frac{1}{12m}} \left( 1 + \frac{3}{\sqrt{m}} + \frac{6}{\pi m} \right) \cdot m.
  \]
  The right hand part of this is a decreasing function of \(m\) and less than \(2\) for \(m \ge M = 3\).
\end{proof}

The proof of our main bound follows by a simple application of \Cref{th:martingale-pac-bayes}.

\begin{proof}[Proof of \Cref{th:main-bound}]
  We set \(M=3\) and
  \[\vec{U}_i(w) =
    \begin{bmatrix}
      \max(\ell(w, Z_i) - \ell^{\star}(Z_{1:i}), 0) \\
      \max(\ell^{\star}(Z_{1:i}) - \ell(w, Z_i), 0) \\
      1 - |\ell^{\star}(Z_{1:i}) - \ell(w, Z_i)| \\
    \end{bmatrix}
  \]
  in \Cref{th:martingale-pac-bayes}, and bound \(\xi(3, m)\) with \Cref{prop:xi-bound}.
\end{proof}

The relaxed version of our bound is given by the following proposition (which holds for non-negative excess losses).
\begin{proposition}\label{prop:relax}
  For any \(\vec{u}, \vec{v} \in \Delta_3\), with \(v_1 > v_2\), if \(\kl(\vec{u} \| \vec{v}) \le b\), then
  \begin{align*}
    v_1 - v_2  &\le \kl^{-1}(u_1\|b) - \kl^{-1}_{\operatorname{LB}}(u_2\|b) \\
        &\le u_1 - u_2 + 2\sqrt{b \cdot (u_1 + u_2)} + 2b,
  \end{align*}
  where \(\kl^{-1}_{\operatorname{LB}}u\|b) := \inf\{ r \in [0, 1] : \kl(u\|r) \le b\}\) is the \emph{lower} tail small-kl inversion.
\end{proposition}

\begin{proof}
  Firstly, we recall \citep[Proposition 9]{adams} that \(\kl(\vec{u} \| \vec{v}) \ge \kl(u_i \| v_i)\) for any \(i\), which immediately gives the first inequality upon inversion.
  The first term is bounded with \(v_1 \le \kl^{-1}(u_1 \| b) \le u_1 + \sqrt{2bu} + 2b\) as in the relaxation of Maurer's bound.
  Next we know that by Taylor's theorem, for any \(0 \le q < p \le 1\), there exists \(u \in [p, q]\) such that
  \[ \kl(q \| p) = \frac{(p-q)^2}{2u(1-u)}.\]
  Thus, if \(p \le \frac12\),
  \[ \kl(q \| p) \ge \frac{(p-q)^2}{2q(1-q)} \ge \frac{(p-q)^2}{2q}.\]
  If \(v_2 < v_1\) then \(v_2 \le \frac12\) and by substitution
  \[
    -v_2 \le -u_2 + \sqrt{2b v_2}.
  \]
  The proof is completed by summing these and applying the bound \(\sqrt{a} + \sqrt{b} \le \sqrt{2(a + b)}\) for \(a, b \ge 0\) (square both sides and subtract \(a+b\) to reduce this to Young's inequality).
\end{proof}

\begin{theorem}[Martingale Chernoff-Hoeffding Inversion]\label{th:martingale-chernoff}
  Let \(U_i \in [0, 1], i = 1, \dots, m\) have conditional expectations \(\E[U_i | U_{1:i-1}] = \mu_i\), and averages \(\bar{U} := \frac{1}{m} \sum_{i=1}^m U_i\), \(\bar{\mu} = \frac{1}{m} \sum_{i=1}^m \mu_i\).
  Then with probability at least \(1{-}\delta\),
  \[
    \bar{\mu} \le \kl^{-1} \left( \bar{U} \middle\| \frac{\log\frac{1}{\delta}}{m} \right).
  \]
\end{theorem}
The proof is deferred to \Cref{appendix:proof}.

\begin{proof}[Proof of \Cref{th:debias-bound}]
  Apply \Cref{th:martingale-chernoff} to the sequence of \(Z_i\)-dependent variables given by \(\E_{W \sim \rho^{\star}_i}\ell[\ell(W, Z_i)]\).
\end{proof}

\section{ASIDE: AN ALTERNATIVE BOUND FOR SIGNED LOSSES}
We note that \Cref{th:martingale-pac-bayes} could instead be used to obtain high-probability upper bounds on a ``signed'' loss function \(\exloss: \mathcal{W} \times \mathcal{Z} \to [-1, 1]\), that is tighter when \(\exloss \approx 0\) on the training set.
In this alternative bound, the only re-definition is of the quantities
\begin{align*}
    \emperr_{+}(\rho) &:= \E_{W \sim \rho} \left[\frac{1}{m} \sum_{i=1}^m \max(\exloss(W, Z_i), 0) \right], \\
    \emperr_{-}(\rho) &:= \E_{W \sim \rho} \left[\frac{1}{m} \sum_{i=1}^m \max(-\exloss(W, Z_i), 0) \right].
\end{align*}
The main bound then holds in an unchanged way.
This approximately relaxes the recent PAC-Bayes split-kl inequality (as applied to signed losses) from \citet{DBLP:journals/corr/abs-2206-00706}, by combining with \Cref{prop:relax}, giving
\begin{equation}\label{eq:split-kl}
  \err \le \kl^{-1}\left(\emperr_+\middle\|\frac{\comp + \lgdelta}{m}\right) - \kl^{-1}_{\operatorname{LB}}\left(\emperr_{-}\middle\|\frac{\comp + \lgdelta}{m}\right).
\end{equation}
This is essentially the same as the above-mentioned bound, except that we have \(\lgdelta = \log\frac{2m}{\delta}\) while in their bound \(\lgdelta = \log\frac{4\sqrt{m}}{\delta}\), so the constants in ours are slightly worse, as in \Cref{section:relax-to-bounded}.
Their main bound is not limited to such signed losses, but is primarily aimed at losses where there are three different special values to be focused on (in the simple case, \(\{-1, 0, 1\}\), as here).
We note that the techniques they use to do this could also be applied using some of our ideas to the bound of \citet{adams}.

\section{EXPERIMENTS}\label{section:experiments}

In this section we empirically compare our bound to that of \citet{maurer} and the Unexpected Bernstein \citep{unexpected}, with a particular focus on the tightening arising through de-biasing by online estimators.

We replicate the experimental setup of \citet{unexpected}, looking which looks at classification with the 0-1 loss by logistic regression of UCI datasets.

The data space is \(\mathcal{Z} = \mathcal{X} \times \mathcal{Y} = \Re^d \times \{0, 1\}\).
Our hypotheses take the form \(h_w(x) = \1_{\psi(w \cdot x) > \frac12}\), where \(\1\) is the indicator function and \(\phi(t) = 1/(1+e^{-t})\) is the standard logistic function.
The 0-1 loss can be written as \[\ell(w, (x, y)) = |y - \1_{\psi(w \cdot x) > \frac12}|.\]
Specifically, we look at learning \(w\) by regularised logistic regression, which (for sample \(S\) and regularisation constant \(\lambda\)) outputs
\[
  \logisticregression_{\lambda}(S) = \operatorname*{argmin}_{w \in \mathcal{W}} \frac{\lambda \|w\|^2}{2} + \frac{1}{|S|} \sum_{(x, y) \in S} \sigma_{w}(x, y)
\]
with
\[
\sigma_w(x, y) = - y \log \psi(w \cdot x) - (1-y) \log(1 - \psi(w \cdot x)).
\]
This is solved empirically using the L-BFGS algorithm \citep{DBLP:journals/mp/LiuN89}.

We set \(\delta{=}0.05\) and \(\lambda{=}0.01\) on all datasets.
In our bounds we choose posterior \(\rho(S) = \mathcal{N}(\logisticregression_{\lambda}(S), \sigma^2 I)\), with \(\sigma^2 \in \{1/2, \dots, 1/2^J : J = \ceil{\log_2 m}\}\) chosen to minimise the bound being considered.
For our prior we fix \(\pi = \mathcal{N}(0, \sigma_0^2 I)\).
Note that we are not using data-dependent priors as originally studied in \citet{unexpected}, in order to isolate the effect of de-biasing; data-dependent PAC-Bayes priors are a rich topic in their own right.

The sequence of online estimators for our bound and the Unexpected Bernstein are chosen as the deterministic predictors outputted by \(\logisticregression_{\lambda}(Z_{1:i-1})\); for computational reasons we update these only after every \(150\) examples, so that each new online estimator predicts the next \(150\) points.
For the first \(150\) data points the online estimators are not yet effective so we simply choose them to have zero error, which does not change the bound on the loss of the online estimators.

The experiments use several UCI datasets, encoded and pre-processed using the same methods as \citet{unexpected}.
Specifically, we encode categorical variables in \(0{-}1\) vectors (increasing the effective dimension of the feature space), remove any instances with missing features, and scale each feature to have values in \([-1, 1]\).
Experiments are repeated \(20\) times with different data shuffling and test-train allocation, and expectation with respect to Gaussian variables are evaluated using Monte Carlo estimates.

\begin{table}
\begin{center}
\begin{tabular}{lcccc}
    \toprule
Dataset   & Test & Maurer & UB    & Ours  \\
    \midrule
Haberman  &  0.273  &  \textbf{0.415}  &  0.583  &  0.501 \\
Breast-C  &  0.037  &  \textbf{0.139}  &  0.208  &  0.164 \\
Tictactoe &  0.043  &  \textbf{0.214}  &  0.369  &  0.245 \\
Banknote  &  0.050  &  \textbf{0.129}  &  0.192  &  0.136 \\
kr-vs-kp  &  0.045  &  0.167  &  0.247  &  \textbf{0.164} \\
Spambase  &  0.169  &  0.324  &  0.501  &  \textbf{0.306} \\
Mushroom  &  0.003  &  \textbf{0.055}  &  0.082  &  0.056 \\
Adult     &  0.170  &  0.234  &  0.384  &  \textbf{0.211} \\
    \bottomrule
\end{tabular}
\end{center}
\caption{Test error of \(\logisticregression_{\lambda}(S)\), and bounds for \(\rho(S)\) with optimised \(\sigma\) as obtained on the UCI datasets listed.
  The datasets are ranked in order of size, from least examples to most.
  The bounds evaluated are Maurer's small-kl bound, the unexpected Bernstein bound and our \Cref{th:main-bound}, with the latter two using online estimators for de-biasing as described.
  }
\label{table1}
\end{table}

\paragraph{Discussion.}
Empirically we observe that our bound more effectively leverages the de-biasing of online estimators than the unexpected Bernstein, providing a tighter numerical guarantee in every case.
On the smaller datasets it is somewhat weaker than Maurer's bound, but it is close to or better than it on the larger datasets.
This arises because when the number of examples is very small, the online estimators are poor surrogates for the final posterior, and the de-biasing term is only weakly correlated with the loss.
On the larger datasets, the de-biasing process is more effective and our bound is the tightest.

\section{SUMMARY}

In \Cref{th:main-bound} we have provided a new PAC-Bayesian bound which can be used alongside an extension of excess losses.
In particular, this extension of the excess loss is able to use information about the difficulty of examples in a pseudo-online fashion, as the learning algorithm passes over the dataset.
This minimises the variance of our generalised excess loss.
Our new bound is able to leverage this reduced variance to obtain tighter overall generalisation bounds and fast rates under broader settings.

By harnessing the power of online estimators and small-kl-based bounds in a new way, we have provided a new direction for numerical and theoretical improvements in PAC-Bayes bounds.
Information about the difficulty of examples is most easily used for stable algorithms in our framework, which links nicely to further ideas like the complimentary use of data-dependent or distribution-dependent priors.


\bibliography{bibliography}

\clearpage
\appendix

\include{appendix}

\end{document}

%% file: appendix.tex
\onecolumn

\section{ADDITIONAL PROOFS AND THEOREMS}\label{appendix:proof}

\subsection{PROOF OF THEOREM \ref{th:martingale-chernoff}}

\begin{proof}
  We show that
  \begin{equation}\label{eq:chernoff-basic-form}
    \Pr\left( \bar{U} \ge \bar{\mu} + t \right) \le e^{- m \cdot \kl(\bar{\mu} + t \| \bar{\mu})}.
  \end{equation}
  From this, the proof of Theorem 4 in \citet{DBLP:journals/corr/abs-2209-05188} implies our theorem statement.
  By Markov's inequality and the convexity of \(t \mapsto e^{ct}\),
  \begin{align*}
    \Pr&\left( \bar{U} \ge \bar{\mu} + t \right) \;
    \le \; \Pr\left( e^{m \lambda \bar{U}} \ge e^{m\lambda (\bar{\mu} + t)}  \right) \\
    &\le e^{-m\lambda (\bar{\mu} + t)} \E \left[ e^{m \lambda \bar{U}} \right]  \\
    &\le e^{-m\lambda (\bar{\mu} + t)} \E \left[ \prod_{i=1}^m e^{\lambda U_i} \right]  \\
    &\le e^{-m\lambda (\bar{\mu} + t)} \E \left[ \prod_{i=1}^m (1 - U_i + U_i e^{\lambda}) \right]  \\
    &\le e^{-m\lambda (\bar{\mu} + t)} \E \left[ \prod_{i=1}^m (1 - \mu_i + \mu_i e^{\lambda}) \right]  \\
  \end{align*}
  where in the final step we have used the same telescoping property of conditional expectations as in the proof of \Cref{th:martingale-vector-convexity}.
  By the arithmetic-geometric mean inequality, the product term is upper bounded by
  \[
    \left( \frac{1}{m} \sum_{i=1}^m (1 - \mu_i + \mu_i e^{\lambda}) \right)^m = (1 - \bar{\mu} + \bar{\mu} e^{\lambda})^m.
  \]
  Substitution shows that the probability above is upper bounded by
  \[
    \left( \frac{1 - \bar{\mu} + \bar{\mu} e^{\lambda}}{e^{\lambda(\bar{\mu} + t)}} \right).
  \]
  Optimising this bound w.r.t. \(\lambda\) gives the form on \Cref{eq:chernoff-basic-form}.
\end{proof}

\subsection{PAC-BAYES UNEXPECTED BERNSTEIN WITH GENERALISED EXCESS LOSS}

In this section we reproduce the following central result of \citet{unexpected} in the form used by our empirical comparison (which uses de-biasing but not informed priors).

\begin{theorem}[PAC-Bayes Unexpected Bernstein Excess Loss]\label{th:unexpected-bernstein-excess}
  For loss \(\ell \in [0, 1]\), for any fixed \(\eta \in (0, 1)\) and prior \(\pi \in \probmeasure(\mathcal{W})\),
  with probability at least \(1{-}\delta\) over the sample \(S\) simultaneously for any \(\rho \in \probmeasure(\mathcal{W})\)
  \[
     \err(\rho) - \emperr(\rho) \le c_{\eta} \widehat{V}(\rho) + \frac{\KL(\rho, \pi) + \log\frac{1}{\delta}}{m\eta}.
  \]
  Here \(c_{\eta} = \frac{\eta + \log(1-\eta)}{\eta}\),
  \begin{align*}
    \err(\rho) &:= \risk(\rho) - \frac{1}{m} \sum_{i=1}^m \E_{Z} \E_{W \sim \rho^{\star}_i}[\ell(W, Z)], \\
    \emperr(\rho) &:= \emprisk(\rho) - \frac{1}{m} \sum_{i=1}^m \E_{W \sim \rho^{\star}_i}[\ell(W, Z)], \\
    \widehat{V}(\rho) &:= \E_{W \sim \rho}\left[\frac{1}{m}\sum_{i=1}^m |\ell(W, Z_i) - \E_{W' \sim \rho^{\star}_i}\ell(W', Z_i)|^2 \right].
  \end{align*}
\end{theorem}

We note that if the online estimators are fixed these quantities reduce to the standard excess risk terms.
In order to prove this result, we first state and prove some intermediate results.

\begin{proposition}[Unexpected Bernstein Lemma; \citealp{unexpected}, Lemma 13]\label{th:unexpected-lemma}

  Let \(U \le 1\) a.s.; then for any \(\eta \in (0, 1)\)
  \[
    \E e^{\eta(\E[U] - U - c_{\eta} U^2)} \le 1
  \]
  where \(c_{\eta}  = \frac{\eta + \log(1-\eta)}{\eta}\).
\end{proposition}

\begin{proof}
  For \(t < 1\), define the decreasing function
  \[
    f(t) = \frac{\log(1-t) + t}{t^2}.
  \]
  Let \(u \le 1\) and \(\eta \in (0, 1)\), so that \(u\eta \le \eta < 1\).
  Since \(f\) is decreasing,
  \begin{align*}
    f(\eta) \le f(u\eta) \; \implies \; \frac{\log(1-\eta) + \eta}{\eta^2} \le \frac{\log(1-u\eta) + u\eta}{(u\eta)^2} \; \implies \; \exp (\eta c_{\eta} u^2 - \eta u) \le 1 - u\eta.
  \end{align*}
  Setting \(u = U\) and taking the expectation, and using \(1-t \le e^{-t}\),
  \[
    \E \exp (\eta c_{\eta} U^2 - \eta U) \le 1 - \E[U]\eta \le e^{-\E[U]},
  \]
  dividing through by the right hand side gives the result.
\end{proof}

We also give the following unexpected-Bernstein counterpart of \Cref{th:martingale-pac-bayes} which can be used to trivially prove the main result.

\begin{theorem}\label{th:martingale-unexpected-bernstein}
  Let \(U_1(w), \dots, U_m(w)\) be a sequence of random bounded functions, valued in \([0, 1]\), such that
  \[\E[U_i(w) | U_{1}(w), \dots, U_{i-1}(w)] = \mu_i(w)\]
  for \(i = 1, \dots, m\) and all \(w \in \mathcal{W}\).
  Define
  \begin{align*}
    \widehat{U}(\rho) &:= \E_{W \sim \rho}\left[\frac{1}{m} \sum_{i=1}^m U_i(W)\right]  &&\text{and}&  \bar{\mu}(\rho) &:= \E_{W \sim \rho}\left[\frac{1}{m} \sum_{i=1}^m \mu_i(W)\right].
  \end{align*}
  For any fixed \(\pi \in \probmeasure(\mathcal{W}), \delta \in (0, 1), \eta \in (0, 1)\), with probability at least \(1{-}\delta\) (over all \(U_i\)), simultaneously for all \(\rho \in \probmeasure(\mathcal{W})\),
  \[
    \bar{\mu}(\rho) - \widehat{U}(\rho) \le \frac{c_{\eta}}{m} \sum_{i=1}^m \E_{W \sim \rho} |U_i(W)|^2 + \frac{\KL(\rho, \pi) + \log\frac{1}{\delta}}{m}
  \]
  where \(c_{\eta}  = \frac{\eta + \log(1-\eta)}{\eta}\).
\end{theorem}

\begin{proof}
  Firstly, we combine \Cref{th:unexpected-lemma} with recursion of conditional expectations to find that
  \begin{align*}
    \E \exp&\left(\eta \sum_{i=1}^m(\mu_i - U_i - c_{\eta} U_i^2) \right) \\
    &\le \E \left[ \prod_{i=1}^m \exp \left(\eta (\mu_i - U_i - c_{\eta} U_i^2)\right) \right]\\
    &\le \E_{U_{1:m-1}} \left[ \prod_{i=1}^{m-1} \exp \left(\eta (\mu_i - U_i - c_{\eta} U_i^2)\right) \, \cdot \, \E_{U_m}[\exp(\eta (\mu_m - U_m - c_{\eta} U_m^2))|U_{1:m-1}] \right] \\
    &\le \E_{U_{1:m-1}} \left[ \prod_{i=1}^{m-1} \exp(\eta (\mu_i - U_i - c_{\eta} U_i^2)) \right] \\
    &\le 1.
  \end{align*}

  Next, as in the proof of \Cref{th:martingale-pac-bayes}, we combine this with Donsker-Varadhan, Markov's inequality and the independence of \(\pi\) from \(U_{1:m}\) to find the following holds with probability at least \(1{-}\delta\) for all \(\rho\):
  \begin{align*}
    \E_{W \sim \rho} &\left[\eta \sum_{i=1}^m(\mu_i(W) - U_i(W) - c_{\eta} U_i(W)^2)  \right] - \KL(\rho, \pi) \\
    &\le \log \E_{W \sim \pi} \left[\eta \sum_{i=1}^m(\mu_i(W) - U_i(W) - c_{\eta} U_i(W)^2)  \right] \\
    &\le \log \E_{U_{1:m}} \E_{W \sim \pi} \left[\eta \sum_{i=1}^m(\mu_i(W) - U_i(W) - c_{\eta} U_i(W)^2)  \right] \\
    &\le \log \frac{1}{\delta} \E_{W \sim \pi}\E_{U_{1:m}}  \left[\eta \sum_{i=1}^m(\mu_i(W) - U_i(W) - c_{\eta} U_i(W)^2)  \right] \\
    &\le \log \frac{1}{\delta}.
  \end{align*}
  Dividing both sides through by \(m\eta\) gives the result.
\end{proof}

\begin{proof}[Proof of \Cref{th:unexpected-bernstein-excess}.]
  In \Cref{th:martingale-unexpected-bernstein}, set
  \[
    U_i(w) = \ell(w, Z_i) - \E_{W \sim \rho^{\star}_i}\ell(W, Z_i)
  \]
  for each \(i\), which is bounded above by \(1\).
\end{proof}

\subsection{RELAXATION OF MARTINGALE CHERNOFF}

In the following, we prove a relaxation of the inverse small kl which leads to a form much more similar to the unexpected Bernstein, and is used later to motivate our experimental setup.

\begin{proposition}\label{prop:relax-inverse-kl}
  For \(0 \le q < 1\) and \(b > 0\),
  \[
    \kl^{-1} \left( q \middle\| b \right) < \inf_{\eta \in (0, 1)} \left[(1 + c_{\eta}) q + \frac{b}{\eta}\right]
  \]
  where \(c_{\eta} := - \frac{\eta + \log(1-\eta)}{\eta}\).
\end{proposition}

In \citet[Proposition 2.1]{germainPACBayesianLearningLinear2009} it is proved that
\begin{proposition}
  For any \(0 \le q \le p < 1\),
  \[ \sup_{C > 0} \left[ C\Phi_{C}(p) - Cq \right] = \kl(q \| p)\]
  where
  \[ \Phi_{C}(p) = -\frac{1}{C}\log(1 - p + pe^{-C}).\]
\end{proposition}

\begin{proof}[Proof of \Cref{prop:relax-inverse-kl}]
  For any \(C > 0\), \(C\Phi_C(p) - Cq \le \kl(q\|p)\), and thus
  \[
    \kl^{-1}(q \| b) = \sup \{ p \in (0, 1) : \kl(q \| p) \le b\} \le \sup \{ p \in (0, 1) : C\Phi_C(p) - Cq \le b\} = \Phi_C^{-1}(q + b/c)
  \]
  with the latter step following by the invertibility of \(\Phi_C\).
  Since \(1 - e^{-t} \le t\) with equality only at \(t = 0\),
  \[
    \Phi_C^{-1}(t) = \frac{1-e^{-Ct}}{1-e^{-C}} \le \frac{Ct}{1-e^{-C}}.
  \]
  As \(b > 0\) and \(q \ge 0\), we have \(q + b/c \ne 0\) and therefore
  \[
    \Phi_C^{-1}(q + b/c) < \frac{Cq + b}{1 - e^{-C}}.
  \]
  Introducing \(\eta = 1 - e^{-C} \in (0, 1)\) (so that \(C = -\log(1-\eta)\)),
  \[
    \frac{Cq + b}{1 - e^{-C}} =  \frac{-\log(1-\eta)}{\eta} q + \frac{b}{\eta} = (1 + c_{\eta}) q + \frac{b}{\eta}.
  \]
  Chaining these results, and taking an infimum of both sides over the free variable \(\eta \in (0, 1)\) completes the proof.
\end{proof}

\section{FULL BOUNDS USED IN EXPERIMENTS}\label{section:full-bounds}

In our experiments, we use the following bounds, obtained by combining (through a union bound) \Cref{th:debias-bound} with \Cref{th:main-bound} or \Cref{th:unexpected-bernstein-excess} with online estimators.
In the unexpected Bernstein case, we combine the result with a grid over possible values of \(\eta\), in the same way as the original paper.
over possible values of \(\eta\).

\begin{theorem}[Generalisation Loss Bound]
  Fix \(\mathcal{D} \in \probmeasure(\mathcal{Z}), \pi \in \probmeasure(\pi), \delta \in (0, 1)\), and \(\ell: \mathcal{Z} \times \mathcal{W} \to [0, 1]\).
  With probability at least \(1-{\delta}\) over \(Z_{1:m} = S \sim \mathcal{D}^m\), for a sequence of online estimators \(\rho^{\star}_i \in \probmeasure(\mathcal{W})\) with \(i=1,\dots, m\), where each depenends only on the \(i-1\) examples \(Z_{1:i-1}\), and for any posterior \(\rho \in \probmeasure(\mathcal{W})\), the following holds
  \[
    \risk(\rho) \le \phi\left(
      \begin{bmatrix}
      \emperr_{+}(\rho) \\
      \emperr_{-}(\rho) \\
      1 - \emperr_{+}(\rho) - \emperr_{-}(\rho)\\
      \end{bmatrix}, \; \frac{\KL(\rho, \pi) + \log\frac{4m}{\delta}}{m} \right) +
       \kl^{-1}\left( \frac{1}{m} \sum_{i=1}^m \E_{W \sim \rho_i^{\star}}[\ell(W, Z_i)] \middle\| \frac{\log\frac{2}{\delta}}{m} \right)
  \]
  with
  \begin{align*}
      \emperr_{+}(\rho) &:= \E_{W \sim \rho} \left[\frac{1}{m} \sum_{i=1}^m \max(\ell(W, Z_i) - \E_{W \sim \rho_i^{\star}}[\ell(W, Z_i)], 0) \right], \\
      \emperr_{-}(\rho) &:= \E_{W \sim \rho} \left[\frac{1}{m} \sum_{i=1}^m \max(\E_{W \sim \rho_i^{\star}}[\ell(W, Z_i)]-\ell(W, Z_i), 0) \right].
  \end{align*}
\end{theorem}

\begin{theorem}[Unexpected Bernstein for Generalisation Loss]
  Fix \(\mathcal{D} \in \probmeasure(\mathcal{Z}), \pi \in \probmeasure(\pi), \delta \in (0, 1)\), and \(\ell: \mathcal{Z} \times \mathcal{W} \to [0, 1]\).
  With probability at least \(1-{\delta}\) over \(Z_{1:m} = S \sim \mathcal{D}^m\), for a sequence of online estimators \(\rho^{\star}_i \in \probmeasure(\mathcal{W})\) with \(i=1,\dots, m\), where each depenends only on the \(i-1\) examples \(Z_{1:i-1}\), and for any posterior \(\rho \in \probmeasure(\mathcal{W})\), the following holds
  \[
    \risk(\rho) \le \emprisk(\rho) + \inf_{\eta \in \mathcal{G}} \left[ c_{\eta} \widehat{V}(\rho) + \frac{\KL(\rho, \pi) + \log\frac{2K}{\delta}}{m\eta} \right]
    + \kl^{-1}\left( \frac{1}{m} \sum_{i=1}^m \E_{W \sim \rho_i^{\star}}[\ell(W, Z_i)] \middle\| \frac{\log\frac{2}{\delta}}{m} \right)
  \]
  where
  \begin{align*}
    \widehat{V}(\rho) &:= \E_{W \sim \rho}\left[\frac{1}{m}\sum_{i=1}^m |\ell(W, Z_i) - \E_{W' \sim \rho^{\star}_i}\ell(W', Z_i)|^2 \right] \\
    \intertext{and}
    \mathcal{G} &:= \left\{\frac12, \dots, \frac{1}{2^K} : K = \ceil*{\log_2 \left( \frac12 \sqrt{\frac{n}{\log(1/\delta)}} \right)}\right\}.
  \end{align*}
\end{theorem}

\begin{proof}
  Combine \Cref{th:unexpected-bernstein-excess} with \Cref{th:debias-bound} and take a union bound over the grid \(\mathcal{G}\).
\end{proof}

\subsection{BOUNDING THE ONLINE ESTIMATOR LOSS}

We note here that \citet{unexpected} originally used an alternative bound to go from the excess loss to the Generalisation risk.
Instead of \Cref{th:debias-bound} as we use above, they used the bound
\[
  \frac{1}{m} \sum_{i=1}^m \E_{Z}\E_{W \sim \rho_i^{\star}}[\ell(W, Z)] \le \frac{1}{m} \sum_{i=1}^m \E_{W \sim \rho_i^{\star}}[\ell(W, Z_i)]  +  \inf_{\eta \in \mathcal{G}} \left[ \frac{c_{\eta}}{m} \sum_{i=1}^m \E_{W \sim \rho^{\star}_i}|\ell(W, Z_i)|^2 + \frac{\KL(\rho, \pi) + \log\frac{|\mathcal{G}|}{\delta}}{m\eta} \right].
\]
In the case of the 0-1 misclassification loss used in our experimental setup, where \(\ell^2 = \ell\), this simplifies to the following:
\[
  \frac{1}{m} \sum_{i=1}^m \E_{Z}\E_{W \sim \rho_i^{\star}}[\ell(W, Z)] \le \inf_{\eta \in \mathcal{G}} \left[ (1 + c_{\eta}) \left(\frac{1}{m}\sum_{i=1}^m \E_{W \sim \rho^{\star}_i}\ell(W, Z_i) \right) + \frac{\KL(\rho, \pi) + \log\frac{|\mathcal{G}|}{\delta}}{m\eta} \right].
\]
As we find through \Cref{prop:relax-inverse-kl}, our \Cref{th:debias-bound} implies that
\[
  \frac{1}{m} \sum_{i=1}^m \E_{Z}\E_{W \sim \rho_i^{\star}}[\ell(W, Z)] < \inf_{\eta \in (0, 1)} \left[ (1 + c_{\eta}) \left(\frac{1}{m}\sum_{i=1}^m \E_{W \sim \rho^{\star}_i}\ell(W, Z_i) \right) + \frac{\KL(\rho, \pi) + \log\frac{1}{\delta}}{m\eta} \right].
\]
which is strictly stronger (for example, our result holds simultaneously over all \(\eta \in (0, 1)\) with no grid size penalty, and even for the optimal \(\eta\) this bound is slacker).
Therefore, our result \Cref{th:debias-bound} represents a significant contribution, that can be leveraged in combination with the original unexpected Bernstein bound to tighten it in the case of 0-1 losses (and it may also give tighter numerical bounds with some other loss functions also).
We note that \Cref{th:debias-bound} can easily be combined with the backwards-forwards dataset split used by \citet{unexpected}.

In order to make the fairest empirical comparison between the effects of de-biasing on our bound versus the unexpected Bernstein, we therefore use our bound in the comparison.

\subsection{CALCULATION OF INVERSE KL \(\phi\)}

Based directly on Proposition 11 in \citet{adams}, we give the following proposition, which can be used to calculate \(\phi(\vec{u}, b)\).

\begin{proposition}
    Fix \(\vec{u} \in \Delta_3\) and \(b > 0\).
    Define the increasing function
    \[
    f_{\vec{u}}(s) := \log \left( u_1 + \frac{u_2}{1 + 2s} + \frac{u_3}{1 + s} \right) + u_2 \log(1+2s)  + u_3 \log(1+s)
    \]
    and its inverse \(s^{\star} := f_{\vec{u}}^{-1}(b)\).
    If \(t := -\exp(-s^{\star}) - 1\), then
    \[
    \phi(\vec{u}, b) = \frac{\frac{u_1}{t+1} - \frac{u_2}{t-1}}{\frac{u_1}{t + 1} + \frac{u_2}{t - 1} + \frac{u_3}{t}}.
    \]
\end{proposition}

Computationally, we can find the inverse \(s^{\star}\) by a simple bisection-search or Newton's method.
Our slight re-parameterisation (where we write \(f\) in terms of \(s\) instead of \(t = -e^{-s} - 1\) as used by \citealp{adams}) of the original result makes this calculation considerably more numerically stable.

We note as an aside that once we have calculated \(s^{\star}\), we can also use it to find the gradients \(\frac{\partial}{\partial u_i} \phi(\vec{u}, b)\) and \(\frac{\partial}{\partial b} \phi(\vec{u}, b)\), which may be useful when directly optimising the bound as an objective.

\section{FURTHER EXPERIMENTAL DETAILS}

Below we provide additional information about the datasets used and tabulated empirical results.

\begin{table}[h]
\begin{center}
\begin{tabular}{lccccc}
    \toprule
Dataset   & Size  & Test & Maurer & UB    & Ours  \\
    \midrule
Haberman  &  306   &  0.2726 \(\pm\) 0.0388  &  0.4140 \(\pm\) 0.0114  &  0.5829 \(\pm\) 0.0176  &  0.5020 \(\pm\) 0.0113  \\
Breast-C  &  699   &  0.0371 \(\pm\) 0.0133  &  0.1387 \(\pm\) 0.0049  &  0.2079 \(\pm\) 0.0070  &  0.1635 \(\pm\) 0.0068  \\
Tictactoe &  958   &  0.0427 \(\pm\) 0.0151  &  0.2148 \(\pm\) 0.0056  &  0.3683 \(\pm\) 0.0215  &  0.2456 \(\pm\) 0.0069  \\
Banknote  &  1372  &  0.0498 \(\pm\) 0.0113  &  0.1292 \(\pm\) 0.0033  &  0.1926 \(\pm\) 0.0075  &  0.1359 \(\pm\) 0.0038  \\
kr-vs-kp  &  3196  &  0.0449 \(\pm\) 0.0084  &  0.1670 \(\pm\) 0.0023  &  0.2466 \(\pm\) 0.0039  &  0.1633 \(\pm\) 0.0029  \\
Spambase  &  4601  &  0.1694 \(\pm\) 0.0132  &  0.3238 \(\pm\) 0.0027  &  0.5015 \(\pm\) 0.0082  &  0.3054 \(\pm\) 0.0032  \\
Mushroom  &  8124  &  0.0026 \(\pm\) 0.0013  &  0.0551 \(\pm\) 0.0007  &  0.0820 \(\pm\) 0.0015  &  0.0565 \(\pm\) 0.0009  \\
Adult     &  32561 &  0.1696 \(\pm\) 0.0045  &  0.2341 \(\pm\) 0.0013  &  0.3842 \(\pm\) 0.0024  &  0.2108 \(\pm\) 0.0014  \\
    \bottomrule
\end{tabular}
\end{center}
\caption{Test error of \(\logisticregression_{\lambda}(S)\), and bounds for \(\rho(S)\) with optimised \(\sigma\) as obtained on the UCI datasets listed.
  The datasets are ranked in order of size, from least examples to most.
  This size (listed) is the dataset size before the \(20\%\) test set is removed.
  The bounds evaluated are Maurer's small-kl bound, the unexpected Bernstein bound and our \Cref{th:main-bound} as described in \Cref{section:full-bounds}, with the latter two using online estimators for de-biasing as in \Cref{section:experiments}.
  Results are an average of \(20\) runs with standard errors provided.
  }
\label{table1}
\end{table}

\vfill